%% file: main.tex
\pdfoutput=1

\documentclass[11pt]{article}

\usepackage{EMNLP2022}

\usepackage{times}
\usepackage{latexsym}
\usepackage[T1]{fontenc}

\usepackage[utf8]{inputenc}

\usepackage{microtype}

\usepackage{inconsolata}

\usepackage{times}
\usepackage{colortbl}
\usepackage{latexsym}
\usepackage{amsmath}
\usepackage{amssymb}
\usepackage{array}
\usepackage{pifont}
\usepackage{microtype}
\usepackage{tabularx}
\usepackage{adjustbox}
\usepackage{enumitem}

\usepackage{multirow}
\usepackage{dsfont}
\usepackage{booktabs}
\usepackage{algorithm2e}
\usepackage{MnSymbol}
\usepackage{scalerel}
\usepackage{mathrsfs}
\usepackage{pifont}

\usepackage[most]{tcolorbox}

\title{Training Language Models with Memory Augmentation}

\author{Zexuan Zhong$^\dagger$ \quad Tao Lei\thanks{$\ \ $TL currently works at Google Research. The collaboration was initialized before TL joined Google.} \quad Danqi Chen$^\dagger$\\
$^\dagger$Princeton University
\\
\texttt{\{zzhong, danqic\}@cs.princeton.edu, taole@google.com}}

\input{header}

\begin{document}
\maketitle

\input{sections/0-abstract}

\input{sections/1-intro}

\input{sections/3-background}

\input{sections/4-method}

\input{sections/5-experiments}

\input{sections/6-analysis}

\input{sections/2-related}

\input{sections/7-conclusion}

\newpage
\section*{Limitations}
We discuss limitations of our research as follows.

\begin{itemize}
    \item Despite the strong performance achieved by our approach when incorporating a large set of external memory, it results in a reduced inference efficiency at the same time due to the nearest neighbor search. For example, the model is $10 \times$ slower when incorporating external memory. This issue can be more crucial when the external memory is even larger.
    Potential solutions to this issue include (1) constructing the memory using a coarser granularity (e.g., text blocks)~\cite{borgeaud2021improving};
    (2) compressing the external memory set and reducing the dimension of memory representations~\cite{he2021efficient}.
    
    \item We mainly experiment with Transformer-based models and additionally adapt our approach to SRU++~\cite{lei2021srupp}. We believe our approach is compatible with other architectures or techniques such as Transformer-XL~\cite{dai2019transformer} and Compressive Transformer~\cite{rae2019compressive}. We plan to explore them as future work.

    \item We evaluate our approach on machine translation to test the generality of $\ours$ to other generation tasks.
    However, due to compute limitation, we only evaluate it on a small dataset (i.e., {\iwslt}), which consists of 4M tokens in the external memory.
    We leave the evaluation on larger machine translation datasets as future work.
    
    \item Our paper mainly studies language modeling tasks and machine translation tasks. Although we believe our approach is compatible with all language generation tasks, how to adapt {\ours} to natural language understanding tasks such as text classification still remains an open question.
    
    \item The biggest model we experimented with consists of 247M parameters due to our compute limit. The state-of-the-art auto-regressive LMs contain hundreds of billions of parameters~\cite{brown2020language}. We hope to see future efforts in scaling up our approach and evaluating the effectiveness on large LMs.
    
\end{itemize}
\section*{Ethical Considerations}
Our proposed approach leverages external memory to achieve strong results on multiple language modeling benchmarks.
In our experiments, we construct the external memory using the corpus on which the model is trained, while it can be constructed using any corpus.
In general, we suggest practitioners constructing external memory using a public corpus, as retrieving from the external datastore can cause information leakage from the corpus.
We acknowledge this ethical consideration and caution those who apply our approach to privacy-sensitive domains.

\section*{Acknowledgments}
We thank Jane Pan, Howard Chen, Alexander Wettig, Tianyu Gao, Kaiyu Yang, Mengzhou Xia, Jinhyuk Lee, and the members of Princeton NLP group for helping with proofreading and providing valuable feedback. This research is partially supported by the James Mi *91 Research Innovation Fund for Data Science  and a gift from Apple. ZZ is also supported by a JP Morgan PhD fellowship.

\bibliography{anthology,custom}
\bibliographystyle{acl_natbib}

\input{appendix}

\end{document}

%% file: header.tex
\DeclareMathOperator{\simfunc}{sim}

\newcommand{\ours}{\textsc{Trime}}
\newcommand{\ourlocal}{\textsc{TrimeLM}}
\newcommand{\ourlong}{\textsc{TrimeLM}_{\text{long}}}
\newcommand{\ourext}{\textsc{TrimeLM}_{\text{ext}}}
\newcommand{\ourmt}{\textsc{TrimeMT}_{\text{ext}}}
\newcommand{\wiki}{\textsc{WikiText-103}}
\newcommand{\enwik}{\textsc{Enwik8}}
\newcommand{\iwslt}{\textsc{IWSLT'14}}
\newcommand{\book}{\textsc{BooksCorpus}}

\newcommand{\trainm}{\mathcal{M}_\text{train}}
\newcommand{\testm}{\mathcal{M}_\text{eval}}
\newcommand{\localm}{\mathcal{M}_\text{local}}
\newcommand{\longm}{\mathcal{M}_\text{long}}
\newcommand{\externalm}{\mathcal{M}_\text{ext}}

\newcommand\ti[1]{\textit{#1}}

\newcommand\tf[1]{\textbf{#1}}

\newcommand{\tableindent}{~~}

\newcommand{\pfix}{\text{\ding{72}}}

\newtcbox{\hlprimarytab}{on line, rounded corners, box align=base, colback=white!10,colframe=white,size=fbox,arc=3pt, before upper=\strut, top=-2pt, bottom=-4pt, left=-2pt, right=-2pt, boxrule=0pt}
\newtcbox{\hlprimarytabg}{on line, rounded corners, box align=base, colback=gray!10,colframe=white,size=fbox,arc=3pt, before upper=\strut, top=-2pt, bottom=-4pt, left=-2pt, right=-2pt, boxrule=0pt}
\newtcbox{\hlsecondarytab}{on line, box align=base, colback=red!10,colframe=white,size=fbox,arc=3pt, before upper=\strut, top=-2pt, bottom=-4pt, left=-2pt, right=-2pt, boxrule=0pt}

\newcommand{\dashifted}{\raisebox{0.5\depth}{\tiny$\downarrow$}}
\newcommand{\uashifted}{\raisebox{0.5\depth}{\tiny$\uparrow$}}
\newcommand{\da}[1]{{\small\hlprimarytab{\dashifted{#1}}}}
\newcommand{\dagray}[1]{{\small\hlprimarytab{\dashifted{#1}}}}
\newcommand{\ua}[1]{{\small\hlprimarytab{\uashifted{#1}}}}
\newcommand{\uagray}[1]{{\small\hlprimarytab{\uashifted{#1}}}}

%% file: sections/0-abstract.tex
\begin{abstract}

Recent work has improved language models (LMs) remarkably by equipping them with a non-parametric memory component.
However, most existing approaches only introduce memories at testing time or represent them using a separately trained encoder, resulting in suboptimal training of the language model.
In this work, we present {\ours}, a novel yet simple training approach designed for training LMs with memory augmentation. Our approach uses a training objective that directly takes in-batch examples as accessible memory. We also present new methods for memory construction and data batching, which are used for adapting to different sets of memories---local, long-term, and external memory---at testing time.
We evaluate  {\ours}  on multiple language modeling and machine translation benchmarks and show that it is able to achieve significant improvements across all the settings. Concretely, {\ours} reduces the perplexity from 18.70 to 15.37 on {\wiki}, by effectively leveraging a large memory set from the training corpus. Compared to standard LM training, {\ours} adds negligible computational overhead and is compatible with different neural architectures, making it a versatile solution for training memory-augmented LMs.\footnote{Our code and pre-trained models are publicly available at \url{https://github.com/princeton-nlp/TRIME}.}

\end{abstract}

%% file: sections/1-intro.tex
\section{Introduction}
\label{sec:intro}

Memory augmentation has become a remarkable approach to enhance language modeling performance without significantly increasing the amount of parameters and computation.
By accessing memory units such as a neural cache of recent inputs~\cite{merity2016pointer,grave2016improving} and an external look-up table~\cite{khandelwal2020generalization}, a memory-augmented language model (LM) enjoys increased memorization capacity and sets new state-of-the-art records in various language modeling benchmarks.

\begin{figure}[!t]
\centering
\includegraphics[width=1.05\linewidth]{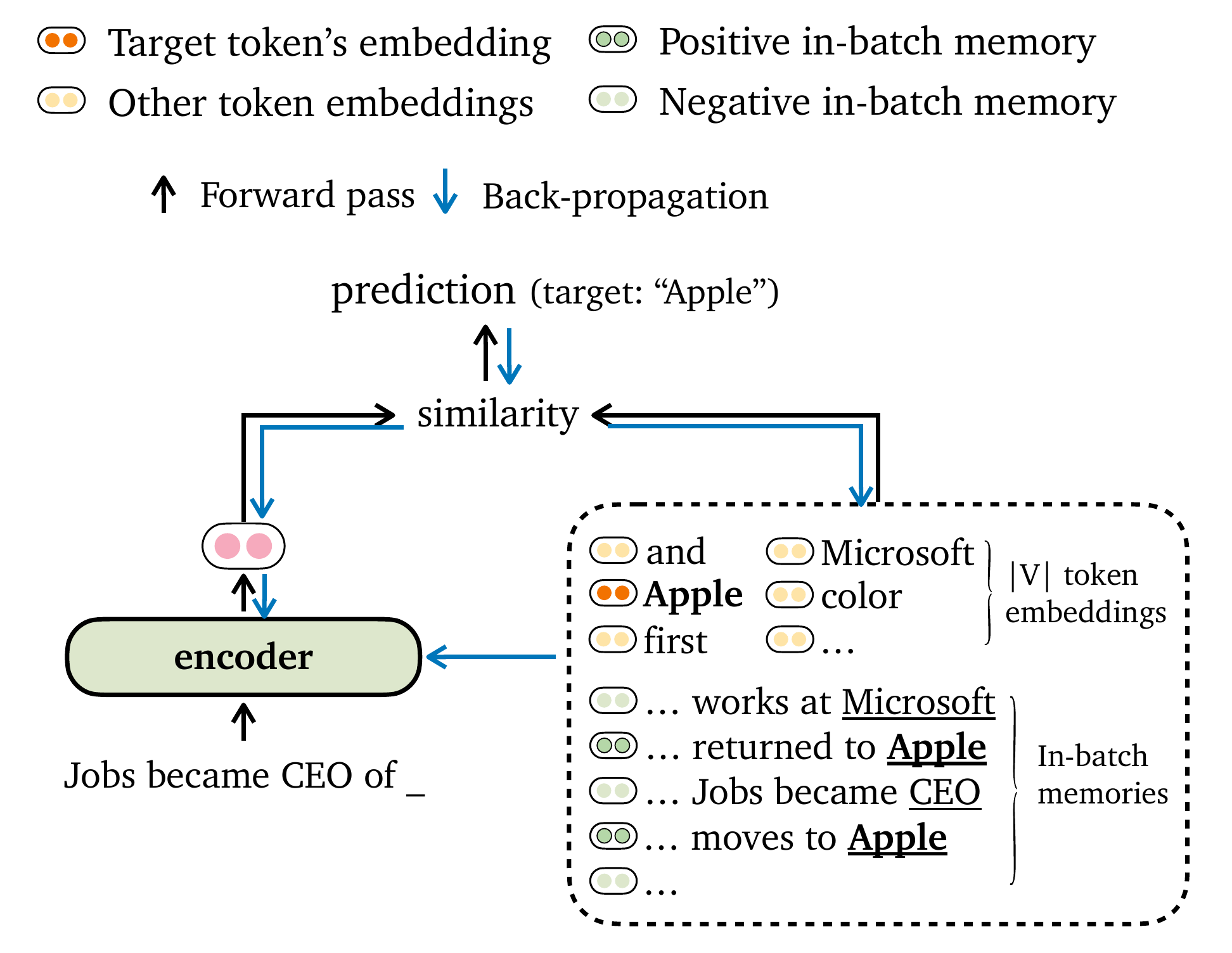}
\caption{An illustration of our training objective. Our objective aligns the hidden representation with both token embeddings and a set of in-batch contextualized representations that are constructed during training. }
\label{fig:training_loss}
\vspace{-1em}
\end{figure}

A major limitation of existing approaches, however, is that the memory units are either introduced at \ti{testing time}~\cite{grave2016improving,grave2017unbounded,khandelwal2020generalization} or taken from a \emph{separately} trained model~\cite{yogatama2021adaptive}.
As a consequence, they are not directly optimized during the training process, resulting in a missed opportunity to achieve even stronger results.
In this paper, we pioneer and present a novel yet simple training approach {\ours} (\tf{Tr}aining with \tf{I}n-batch \tf{M}emories)\footnote{We can also interpret {\ours} as \ti{three} types of \ti{memories}, as we will elaborate in the paper.}, that is well-suited for memory augmentation in language modeling.
Our approach makes two major departures compared to standard language model training:

\paragraph{Training objective}
Inspired by contrastive representation learning, we propose a training objective that directly leverages in-batch examples as accessible memory (Figure~\ref{fig:training_loss}).
Our training objective is closely connected to neural cache models~\cite{grave2016improving,merity2016pointer} and nearest-neighbor language models~\cite{khandelwal2020generalization}, where the next-token probabilities are calculated by comparing encoder outputs against static token embeddings \emph{and} memory representations.
However, previous work only considers incorporating memories at testing time, while we do for both training and testing.

\paragraph{In-batch memory construction}

With this training objective in mind, the key challenge is how to construct memories effectively \ti{during training} while keeping it efficient.
We identify three types of memories that can be leveraged at testing time and have been explored in the literature: (a) \ti{local} memory denotes the words that appear in the recent past and are modeled using attention~\cite{vaswani2017attention}; (b) \ti{long-term} memory\footnote{Long-term memory may have different interpretations in other contexts and we use \ti{long-term} memory to refer to long-range context in modeling long sequences, following previous work~\cite{martins2022infty,wu2022memorizing}.} denotes long-range context from the same document but cannot be directly accessed due to the limit of input length; (c) \ti{external} memory is used to store the entire training set or any additional corpus~\cite{khandelwal2020generalization,borgeaud2021improving}. %

To better leverage these memories at testing time, we devise new \ti{data batching} strategies to improve the construction of training memories (\S\ref{sec:adaption}).
By packing consecutive segments from the same document in one training batch, our model can access long-term memories beyond the attention context. We pack segments from other documents that have high lexical overlap as a proxy to all external memory units.
Importantly, these working memories are generated on the fly during training, allowing us to back-propagate to all memory representations.

\input{tables/comparison.tex}

We instantiate {\ours} in three models by considering different sets of training and testing memories (Table~\ref{tab:comparison}) and evaluate them on multiple language modeling and machine translation benchmarks. We highlight our results as follows:
\begin{itemize}[nosep,wide,labelindent=0.4cm,leftmargin=0cm]
    \item We first show that we can simply optimize a language model using our training objective \ti{without} long-term and external memory. Without any other modifications, we demonstrate that a 247M Transformer-based model can achieve an improved perplexity from 18.70 to 17.76 on {\wiki}~\cite{merity2016pointer} with negligible overhead. This model can be viewed as {a simple replacement} for vanilla language models.
    \item By training with consecutive segments in the same batch, our approach is capable of leveraging very \emph{long context} at testing time---up to 15k-25k tokens on {\wiki} and {\enwik}~\cite{enwik8}. Our approach achieves at least competitive performance as previous works~\cite{dai2019transformer,martins2022infty,ji2022lamemo} that modify the Transformer architecture to incorporate memories from previous segments, yet our solution is conceptually simpler and computationally cheaper.
    \item Finally, we train language models by incorporating all other segments in the same batch as memories. Our model works better with a \emph{large datastore} at testing time and improves over the kNN-LM model~\cite{khandelwal2020generalization} by reducing the test perplexity from 16.23 to 15.41 on {\wiki}. We also demonstrate significant improvements over the kNN-MT baseline~\cite{khandelwal2021nearest} on an {\iwslt} De-En machine translation task.
\end{itemize}
\vspace{-1em}
In summary,
we propose a simple approach {\ours} for optimizing language models with memory augmentation and demonstrate consistent and significant gains in multiple experimental settings.
Our approach only uses memories at the final prediction step, and hence adds little computational overhead and can be combined with different model architectures such as recurrent networks and other attention variants~\cite{lei2021srupp,dai2019transformer,rae2019compressive}.
We hope that our work can encourage the research community to think about better training objectives for language models, given their significant societal impacts~\cite{brown2020language,chowdhery2022palm,zhang2022opt}.

%% file: tables/comparison.tex
\begin{table}[t]
    \centering
    \resizebox{0.94\columnwidth}{!}{
    \begin{tabular}{lcl}
    \toprule
    & \tf{Training} & \tf{Testing} \\
    & \tf{Memory} & \tf{Memory} \\
    \midrule
    vanilla LM & None & None \\
    cont. cache & None & $\localm$ or $\longm$\\
    kNN-LM & None & $\externalm$ \\
    \midrule
    $\ourlocal$ & $\localm$  & $\localm$ \\
    $\ourlong$ & \S\ref{sec:long_term} & $\localm$, $\longm$ \\
    $\ourext$ & \S\ref{sec:external} & $\localm$, $\longm$, $\externalm$\\
    \bottomrule
    \end{tabular}
    }
    \caption{A comparison between our {\ours} language models and previous approaches: vanilla LM, continuous cache~\cite{grave2016improving,grave2017unbounded}, kNN-LM~\cite{khandelwal2020generalization}. ${\localm}$, ${\longm}$, ${\externalm}$ denote local, long-term and external memories respectively (\S\ref{sec:memory_augmentation}).
    \vspace{-1em}
    }
    \label{tab:comparison}
\end{table}

%% file: sections/3-background.tex
\section{Preliminaries}
\label{sec:preliminary}

\subsection{Language Modeling}
\label{sec:language_modeling}
In this paper, we mainly focus on improving language models, although our solutions may extend to most text generation tasks (see one example of machine translation in \S\ref{sec:machine_translation}). 
Neural language models take a sequence of tokens as context $c_t = x_1, \dots, x_{t-1}$ and map it to a vector representation $f_{\theta}(c_t) \in \mathbb{R}^d$, where $f_{\theta}(\cdot)$ is parameterized by a neural network. The next-token probability is:
\begin{align}
\label{equ:ori_loss}
    P(w \mid c_t) \propto \exp(E^\top_wf_{\theta}(c_t)),
\end{align}
where $E_w \in \mathbb{R}^{d}$ denotes the output embedding of token $w \in \mathcal{V}$.
The parameters are optimized to minimize the negative log-likelihood of ground truth $x_t$ during training.

\subsection{Memory Augmentation}
\label{sec:memory_augmentation}
We consider memory as a set of context-target pairs $\{(c_i, x_i)\}$ following \citet{grave2016improving,khandelwal2020generalization}.
These context-target pairs can be aggregated to obtain the next-token probability weighted by the similarity between hidden representations.\footnote{Other memory-augmented models differ in when the memory was introduced, such as using them in attention, and retrieve texts of different granularity as memory~\cite{guu2020realm,borgeaud2021improving}. }
We formalize three types of context-target memories as follows:

\paragraph{Local memory} The local memory is simply the preceding tokens in the same input. Specifically, for $c_t = x_1, \ldots, x_{t - 1}$, it is defined as:
\begin{align}
\label{eq:local_memory}
\localm(c_t) = \{(c_j, x_j)\}_{1\leq j \leq t-1}.
\end{align}
\citet{grave2016improving} use the local memory at testing time, denoted by the ``continuous cache'' model. However, it has been argued less effective for Transformer-based models because they can already learn to leverage recent tokens in the self-attention layers~\cite{khandelwal2020generalization}. Interestingly, we show that using local memory is still beneficial if we consider it during training.

\paragraph{Long-term memory}
Long-term memory denotes long-range context from the same document, but they cannot be directly accessed by attention. For example, if a document contains 10K tokens, only a short segment of text (e.g., 100-3K tokens) can be fed into a Transformer model because the complexity scales quadratically with the input length.
Formally, we divide a document into consecutive segments $s^{(1)}, \dots, s^{(T)}$, where a segment $s^{(i)}$ contains $L$ contexts $s^{(i)}=\{c_1^{(i)}, \dots, c_L^{(i)}\}$. The long-term memory for $c_{t}^{(i)}$ is:
\begin{align}
\label{eq:long_memory}
\longm(c_{t}^{(i)}) = \{(c_j^{(k)}, x_j^{(k)})\}_{1 \leq k < i, 1 \leq j \leq L}.
\end{align}
 Previous works~\cite{dai2019transformer,rae2019compressive,martins2022infty,ji2022lamemo,wu2022memorizing,lei2021srupp} leverage hidden representations from previous segments with modified Transformer architectures to learn long-range dependency. Our approach does not modify the model architecture and is compatible with these neural architectures.\footnote{ Note that continuous cache  can be naturally extended to long-term memory, as we will experiment later. The earlier continuous cache work was applied to LSTMs on long sequences, as LSTMs can linearly scale with long sequences and there is no need to segment documents.}

\paragraph{External memory} Finally, external memory assumes a large corpus $\mathcal{D}$ and the external memory set can be defined as:
\vspace{-0.5em}
\begin{align}
\label{eq:external_memory}
\externalm = \{(c_j, x_j) \in \mathcal{D}\}.
\end{align}
$\mathcal{D}$ can be simply the training corpus, or a domain-specific corpus when the testing domain shifts (\S\ref{sec:domain}). Note that $|\externalm|$ is usually several orders of magnitude larger than previous two types (e.g.,~$10^8$); accessing all the memories is computationally expensive and requires approximate nearest neighbor search~\cite{johnson2019billion}.

%% file: sections/4-method.tex
\begin{figure*}[!h]
\centering
\vspace{-2em}
\includegraphics[width=0.83\linewidth]{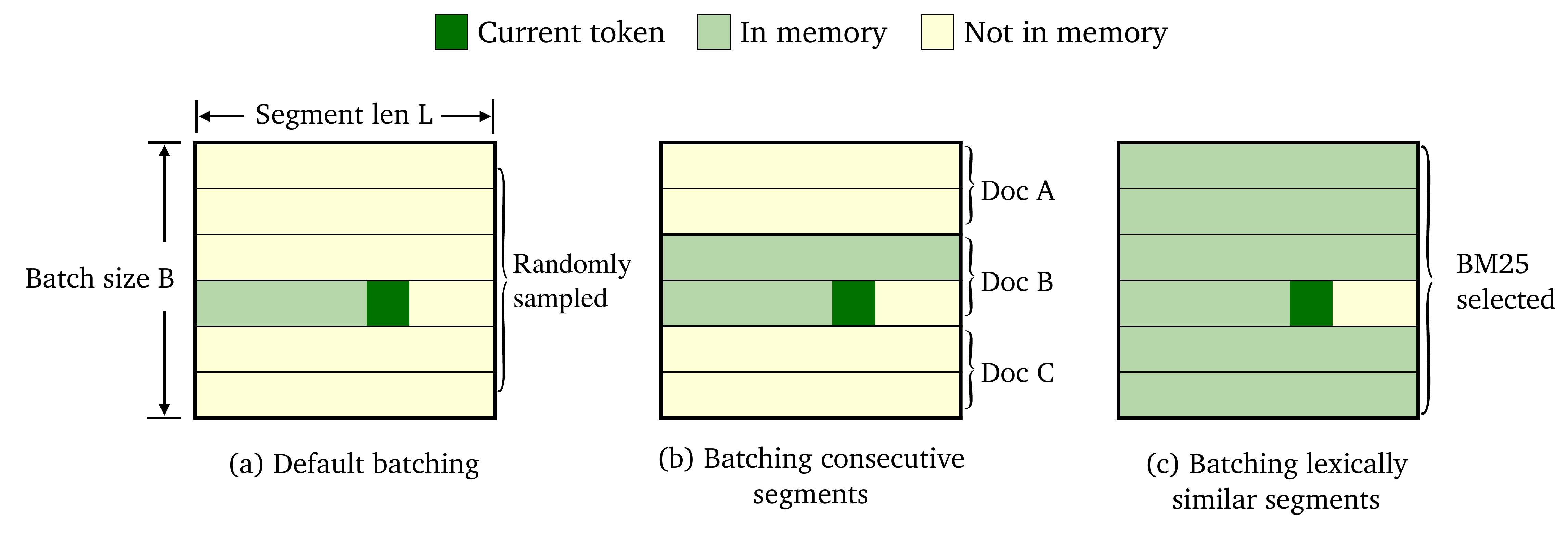}
\vspace{-1em}
\caption{
We present several data batching methods and memory construction strategies, in order to adapt to different sets of testing memories. (a) default batching: all the segments are randomly drawn from the training corpus (\S\ref{sec:local}); (b) we batch consecutive segments from the same document ($m > 1$) in one training batch to better leverage long-range contexts (\S\ref{sec:long_term}); (c) we batch lexically-similar segments in one training batch selected by BM25 to better incorporate a large datastore at testing time (\S\ref{sec:external}).
}
\label{fig:memo_construct}
\end{figure*}

\section{Training with In-batch Memories}
\label{sec:training_objective}
In this section, we propose a new training approach {\ours} for language model training.
Compared to standard language model training, our training objective assumes a set of \ti{training memories} $\trainm = \{({c}_j, {x}_j)\}$. We differentiate training memories from \ti{testing memories}, as they are constructed on the fly during training and may deviate from the testing memories used during inference.  Importantly, the training memories are constructed from the \ti{same} training batch, which enables back-propagating the training signal to the current hidden representation as well as all the memory representations.
We will discuss how to construct training memories in the next section (\S\ref{sec:adaption}) and only discuss the training objective in a general form.

Our training objective is illustrated in Figure~\ref{fig:training_loss}. Given a memory set $\mathcal{M}$ and a context $c$, {\ours} defines the next-token probability distribution as:
\begin{align}
\begin{split}
\label{equ:loss}
    & P(w \mid c)  \propto \exp(E^{\top}_w f_{\theta}(c)) + \\
    & \sum_{(c_j, x_j) \in \trainm: x_j = w} \exp(\simfunc(g_{\theta}(c), g_{\theta}(c_j))).
\end{split}
\end{align}
Here, $f_{\theta}(c)$ is the output representation of a Transformer model and $E_w$ is the token embedding.
$g_{\theta}(\cdot)$ denotes the representations that can be used to compute similarity between $c$ and all the contexts $c_j$ in the memory $\trainm$. It is possible to simply take $g_{\theta} = f_{\theta}$; however, we find that taking $g_{\theta}$ to be the input of the final feed-forward layer in Transformer works better, which is consistent with the observation in \citet{khandelwal2020generalization}.
In addition, $\simfunc(\cdot, \cdot)$ is a similarity function and we found using the scaled dot-product $\simfunc(q, k) = \frac{q \cdot k}{\sqrt{d}}$~\cite{vaswani2017attention} leads to stable training and better performance in our preliminary experiments. 

This training objective can be viewed as a contrastive loss~\cite{hadsell2006dimensionality}: for a context-target pair $(c, w^*)$,
the goal is to align the query representation $f_{\theta}(c)$ (and $g_{\theta}(c)$) with the \ti{static} token representation $E_{w^*}$, {and} \ti{contextualized} representations that share the same next token i.e., $g_{\theta}(c_j)$ for $x_j = w^*$.
Our objective handles rare words nicely---if $w^*$ does not appear in the training memory, the objective will fall back to aligning $f_{\theta}(c)$ with only the word embedding $E_{w^*}$.  Similar to the vanilla training loss (Eq.~\ref{equ:ori_loss}), our {\ours} loss is optimized to minimize the negative log-likelihood of next token $w^*$ and all the parameters $\theta$ and $E_w$ are updated during training.

Our training objective is also inspired by the success of contrastive learning in dense retrieval~\cite{karpukhin2020dense}. As we will show in \S\ref{sec:analysis}, it can help improve retrieving contexts that share the same next token effectively when the set of testing memories is large.
Our objective is also closely connected to the objective used in \citet{grave2016improving,khandelwal2020generalization}, which linearly interpolates the distribution of standard language modeling, and a distribution defined by cache/external datastore, e.g., $P(w \mid c) = (1-\lambda) P_{\text{lm}}(w \mid c) + \lambda P_{\text{kNN}}(w \mid c)$. Our work differs from previous works that we use this objective during \ti{training} (and testing), while they only used it \ti{at testing time}---the key is how to construct training memories that we will elaborate next.\footnote{\citet{grave2016improving} described a ``global normalization'' variant in the paper, which is similar to our objective. However, they only used it at testing time and only considered short-term contexts in calculating the distribution. Other works \cite{merity2016pointer,see2017get} trained a pointer network with a learned gating component for the interpolation---we attempted training with a similar objective earlier and found it to perform worse than our current objective. }
\input{tables/wiki_247M}

\section{Adaption to Different Memories}
\label{sec:adaption}

\paragraph{Inference}
We are interested in incorporating the three types of memories defined in \S\ref{sec:memory_augmentation} and their combinations at testing time.
The testing objective is basically the same as the training objective (Eq.~\ref{equ:loss}) except that we take testing memories as a combination of $\localm$, $\longm$ and $\externalm$.
As $\externalm$ can be very large, we approximate it by retrieving the top-K closest terms to $g_\theta(c)$.
We tune a temperature term $\tau$ to adjust the weight of the memory component (see Appendix~\ref{sec:inference} for details).

\paragraph{Notation}
Throughout this section, we use $L$ to denote segment length, $B$ to denote the total number of segments used in the one training batch, and $m$ to denote the number of consecutive segments from each document in the batch.
Correspondingly, each batch will contain $b\approx \frac{B}{m}$ different documents.
$L$, $B$ and $m$ are hyper-parameters that we will choose for training, and will vary as we consider different memories during inference.

A key challenge is that the testing memories can be very large (e.g., $|\longm| \sim 10^4$ and $|\externalm| \sim 10^8$ in our experiments) and it is computationally infeasible to keep training memories the same as testing memories. 
In the following, we will discuss three ways of constructing training memories and data batching, aiming to reduce the discrepancy between training and testing.
Along the way, we will also present three major model instantiations: ${\ourlocal}$, ${\ourlong}$, ${\ourext}$ (Table~\ref{tab:comparison}), which combine the training strategies and different sets of testing memories.

\subsection{Local Memory}
\label{sec:local}
$\localm$ only considers all the previous tokens in the same segment. It is straightforward that we can simply use $\trainm = \localm$. As shown in Fig.~\ref{fig:memo_construct}(a), we basically do not need to make \ti{any} modifications compared to standard language model training. All we need is to replace the training objective of Eq.~\ref{equ:ori_loss} by our objective in Eq.~\ref{equ:loss}, by incorporating $(c_j, x_j)$, $\forall j < t$ in the memory during both training and testing.  The computational overhead is also negligible compared to running neural encoders on the segment $x_1, \ldots, x_L$ itself. We denote this model as $\ourlocal$, which can be viewed as a lightweight replacement for vanilla language models.  As we will show in the experiments, simply incorporating local memory provides a notable gain on multiple LM benchmarks, showing the effectiveness of training with memories explicitly.

\subsection{Long-term Memory}
\label{sec:long_term}
In order to enable long-term memory augmentation, we pack multiple consecutive segments from the same document in a training batch (i.e., $m > 1$).
For a context-target pair $(c, w)$ in the training batch, its accessible memory $\trainm$ includes tokens from previous segments as well as the preceding tokens in the same segment.
Figure~\ref{fig:memo_construct}(b) illustrates the training batch construction and the training memory for a given token. At testing time, we can use a much longer context: we simply enumerate the number of segments used in $\testm$ and choose the optimum based on the development set.

We denote this model as $\ourlong$. It shares a similar motivation with many previous works which aim to leverage memory from previous segments through attention recurrence~\cite{dai2019transformer,ji2022lamemo}, or memory compression~\cite{rae2019compressive,martins2022infty,wu2022memorizing}.
However, our solution deviates significantly from previous approaches.
First, previous works need to store the hidden representations (of every layer) from previous segments and modify the self-attention layers to incorporate them.
Our approach does not modify the architecture and only uses the outputs from the last layer.
Additionally, previous works use stale memory representations and do not back-propagate gradients to the representations of previous segments, whereas our batching method enables gradient propagation to the memory and previous segments.\footnote{We also attempted using segments in previous training batches as stale representations and did not find any improvement in preliminary experiments.}
As we will show in the experiments, our approach is competitive with previous works while being conceptually simpler and computationally cheaper.

\subsection{External Memory}
\label{sec:external}
Finally, we consider external memory $\externalm$. Since $\externalm$ contains the context-target pairs in a large corpus such as the entire training set, we need to retrieve top-$K$ pairs from $\externalm$ measured by $\simfunc(g_{\theta}(c), g_{\theta}(c_j))$ through (approximate) similarity search (more details are given in \S\ref{sec:exp_details}).

Since the retrieved contexts at testing time are expected to be similar to the query context, we propose a simple heuristic for constructing training memories $\trainm$ by packing segments that have large lexical overlap into the same batch using BM25 scores~\cite{robertson2009probabilistic}.
Specifically, we start with a single segment and repeatedly add segments with highest BM25 scores into the same batch (Appendix~\ref{sec:bm25_batching}).
A high BM25 score indicates that two segments have high lexical overlap and can serve as a good proxy to nearest neighbors in the external memory, which improves our model predictions at testing time.
$\trainm$ contains all tokens from other segments as well as the previous tokens in the same segment (Figure~\ref{fig:memo_construct}(c)). We set $m=1$ during training as many segments from the same document tend to have high lexical overlap and denote this model by $\ourext$.

In practice, when considering tokens from both the current segment and other segments in the batch, we observe that the model tends to leverage local memory more and ignore other segments. To encourage the use of information from other segments, we exclude the local memory from $\trainm$ with a probability of $p$ during training (we find that $p = 90\%$ works the best, see Appendix~\ref{sec:tuning_p}).
This significantly improves performance when the model is evaluated with a large set of external memory.

%% file: tables/wiki_247M.tex
\begin{table*}[ht]
\centering
    \resizebox{1.7\columnwidth}{!}{
\begin{tabular}{lllll}
\toprule
\textbf{Model}                          & \textbf{\#Params} & \textbf{Dev} ($\downarrow$)   & \textbf{Test} ($\downarrow$)  & \textbf{Speed} ($\uparrow$) \\
\midrule
Transformer~\cite{baevski2018adaptive} & 247M & 17.96 & 18.65 &  - \\
~~~~ + continuous cache~\cite{grave2016improving} & 247M & 17.67 & 18.27 & -\\
Transformer-XL~\cite{dai2019transformer} & 257M & - & 18.30 &  - \\
\midrule
Transformer (our run) & 247M & 18.04 & 18.70 & 3.6k t/s\\
~~~~ + continuous cache & 247M & 17.65 & 18.26 \da{0.44} &  3.6k t/s\\
{\pfix} $\ourlocal$ & 247M & 17.10 & 17.76 \dagray{0.94} & 3.6k t/s\\
{\pfix}  $\ourlong$ & 247M & \tf{17.01} & \tf{17.64} {\dagray{1.06}} & 3.6k t/s\\
\midrule
kNN-LM (our run) & 247M & 16.40 & 16.37 & 300 t/s\\
~~~~ + continuous cache & 247M & 16.23 & 16.23  \da{0.14} & 300 t/s\\
{\pfix}  $\ourext$ (w/o $\longm$) & 247M & 15.62 & 15.55 \dagray{0.82} & 300 t/s \\
{\pfix} $\ourext$ & 247M & \tf{15.51} & \tf{15.41} {\dagray{0.94}} & 300 t/s\\
\midrule
kNN-LM~\cite{khandelwal2020generalization}$^\dagger$  & 247M  & 16.06  &   16.12   &    50 t/s   \\
~~~~ + continuous cache~\cite{grave2016improving}$^\dagger$   & 247M   &  15.81 &     15.79 \da{0.33} & 50 t/s \\
{\pfix}  $\ourext^\dagger$ & 247M & \tf{15.40} & \tf{15.37} {\dagray{0.75}} & 50 t/s \\
\bottomrule
\end{tabular}
}
\vspace{-0.5em}
\caption{Performance of our {\ours} models on {\wiki} (247M models, $L =$ 3,072). $^\dagger$: the results are based on computing actual distances instead of using approximated distances returned by FAISS indexes, which requires a large SSD storage. To measure the speed of models (tokens/second), we run the model with a single NVIDIA RTX 3090 GPU and run the FAISS indexer with 32 CPUs.
}
\label{tab:wiki}

\end{table*}

%% file: sections/5-experiments.tex
\section{Experiments}
\label{sec:experiments}

\subsection{Datasets and Tasks}
\label{sec:setup}

We evaluate our approach on two popular language modeling benchmarks: {\wiki}~\cite{merity2016pointer}, {\enwik}~\cite{enwik8}, and a machine translation benchmark: {\iwslt} \textsc De-En. We also evaluate domain-adaptation performance on the {\book} dataset~\cite{zhu2015aligning}.

\tf{{\wiki}} is a word-level language modeling dataset consisting of 103M training tokens. 
We evaluate on two model configurations: one uses a 247M Transformer model and a segment length $L=3,072$ and another one uses a 150M Transformer model with a segment length $L=150$.

\tf{{\enwik}} is a character-level language modeling dataset that contains a total of 100M characters. 
We use a 12-layer Transformer model with a hidden dimension $512$ and segment length $L = 512$.

\tf{{\book}} is a word-level language modeling dataset. We build our own train/dev/test splits which consist of 100M/250K/250K tokens. On this dataset, we evaluate the models trained on {\wiki} to study how our approach can adapt to new domain without re-training.

\tf{\iwslt} \textsc De-En is a machine translation task, which consists of 170K translation pairs. We use a Transformer encoder-decoder model. See Appendix~\ref{sec:dataset} for how we adapt our approach to the machine translation task.

See Appendix~\ref{sec:dataset} for data statistics and task setups and Appendix~\ref{sec:model_config} for model configurations.

\subsection{Training and Inference Details}
\label{sec:exp_details}

We implement our approach using the Fairseq library~\cite{ott2019fairseq}. For $\ourlong$ and $\ourext$, we tune the number of segments used in $\longm$ on the development set during evaluation. Our $\ourext$ model requires building a large datastore at testing time and we use the FAISS library~\cite{johnson2019billion} for approximate nearest neighbor search (details in Appendix~\ref{sec:model_config}).

We first train our model with the standard LM objective (Eq.~\ref{equ:ori_loss}) for the first 5\% updates.
Without this warmup stage, we observe the training process to be unstable probably due to a large variance in the estimated distributions.
We use different memories when evaluating different instantiations of {\ours}, as shown in Table~\ref{tab:comparison}.
We find that when a large set of external memory $\externalm$ is considered during inference, the performance can be improved by linearly interpolating the output distribution and a distribution over the memory, similarly to kNN-LM~\cite{khandelwal2020generalization}.
Thus, we apply an additional linear interpolation to our output probability distribution when considering external memory $\externalm$ (see Appendix~\ref{sec:interpolation} for details).

\subsection{Results: Language Modeling}

\paragraph{{\ourlocal} vs. vanilla LM} We first compare our {\ourlocal} model which only uses local memory during training and testing. Table~\ref{tab:wiki} shows that adding a continuous cache during inference can improve the performance of vanilla Transformer from 18.70 to 18.26, and our {\ourlocal} further improves the perplexity to 17.76. These results suggest that even though the attention mechanism can ``see'' local context, using local memory during both training and testing can still improve model performance. {\ourlocal} has no computational overhead compared to vanilla LM (indicated by the ``speed'' column), making it a simple and better replacement for vanilla language models. Similar trends can be observed in Table~\ref{tab:wiki_150} and Table~\ref{tab:enwik8} (25.87 vs. 25.60 and 1.16 vs. 1.12).
The improvement is much smaller though, due to a much smaller segment length $L$.
More analysis is given in Appendix~\ref{sec:eff_local}.

\input{tables/wiki_150M}

\input{tables/enwik8}

\vspace{-0.5em}
\paragraph{$\ourlong$ leverages long contexts} We then examine our $\ourlong$ model which is trained with the data batching method described in \S\ref{sec:long_term}. As shown in Table~\ref{tab:wiki_150} and Table~\ref{tab:enwik8}, $\ourlong$ improves vanilla Transformer models substantially (i.e., $25.87 \rightarrow 22.66$ on {\wiki} and $1.16 \rightarrow 1.05$ on {\enwik}) by leveraging long-range contexts at inference time.
We find the model achieves its best results when leveraging 15,000 tokens on {\wiki} and 24,576 tokens on {\enwik}, even though the segments used during training are much shorter ($L = $ 150 and 512 respectively).
We also add continuous cache to the vanilla Transformer model and find it to underperform our model, demonstrating the importance of joint training using our approach.

Compared to previous methods which explicitly leverage hidden representations from previous segments~\cite{dai2019transformer,rae2019compressive,martins2022infty,ji2022lamemo,lei2021srupp}, our approach achieves better or at least competitive performance. Different from these approaches which need to store all the hidden representations of every layer and modify the model architecture, we only incorporate the outputs from the last layer---requiring less computations and GPU memory. 
Our approach is orthogonal and can be applied on top of these models. 
To verify this, we adapt our approach to SRU++~\cite{lei2021srupp} (see details in Appendix~\ref{sec:apply_to_sru}). As shown in the bottom block of Table~\ref{tab:enwik8}, $\ourlong$ gains consistently improvement over vanilla SRU++, outperforming previously reported results given the same model size.

\vspace{-0.5em}
\paragraph{$\ourext$ vs. kNN-LM} 
Finally, our model $\ourext$ outperforms the kNN-LM model~\cite{khandelwal2020generalization}, which uses external memory only at testing time---improving the perplexity from 16.23 to 15.41 on {\wiki} (Table~\ref{tab:wiki}). We also evaluate a model which does not use long-term memory (denoted by $\ourext$ w/o $\longm$) for a fair comparison with kNN-LM with continuous cache and the difference is very small (15.55 vs 15.41). Our results suggest that by using contrastive loss and BM25 batching (\S\ref{sec:external}), the model learns to better retrieve and leverage information from a large external memory.

\vspace{-0.5em}
\paragraph{Domain adaptation}
\label{sec:domain}
\input{tables/bookcorpus}

We evaluate the domain-adaptation performance of $\ours$ on {\book}~\cite{zhu2015aligning}.
We take models that are trained on {\wiki} and evaluate them on {\book} without any re-training or fine-tuning.
As shown in Table~\ref{tab:bookcorpus}, a vanilla Transformer model trained on {\wiki} performs poorly on {\book}.
$\ourlocal$ and $\ourlong$ can significantly improve the performance as they leverage local or long-term memory to adapt to the new domain.
By building the external memory using {\book}, both kNN-LM and $\ourext$ perform much better on {\book} compared to the vanilla Transformer model.
$\ourext$ outperforms kNN-LM on domain adaptation. This indicates that although the memory representations are optimized on one domain, our approach does not overfit, and
building an external memory using the target domain dataset enables the model to perform well with domain shifts.

\vspace{-0.3em}
\subsection{Results: Machine Translation}
\label{sec:machine_translation}

To showcase the generality of our training approach {\ours} to other generation tasks, we evaluate our approach on the {\iwslt} de-en translation task. Since it is a sentence-level task, we do not use any local or long-term memory ($\localm$, $\longm$), as there are few repetitive tokens. We denote our model as ${\ours}\text{MT}_{\text{ext}}$.

\input{tables/iwslt}

As shown in Table~\ref{tab:iwslt}, our approach improves the vanilla Transformer by $1.15$ BLEU score and outperforms kNN-MT~\cite{khandelwal2021nearest}.
This demonstrates that our approach is able to improve the performance on other language generation tasks with different memory access.

%% file: tables/wiki_150M.tex
\begin{table}[t]
    \centering

    \resizebox{0.9\columnwidth}{!}{
    \begin{tabular}{lll}
    \toprule
    \textbf{Model} & \textbf{Dev} ($\downarrow$) & \textbf{Test} ($\downarrow$) \\
    \midrule
    Transformer & 28.11 & 29.14  \\
    Transformer-XL & 23.42 & 24.56  \\
    Compressive Transformer & - & 24.41\\
    $\infty$-former & - & 24.22 \\
    LaMemo & 22.98 & 23.77  \\
    \midrule
    Transformer (our run) & 25.31 & 25.87 \\
    ~~~~+ continuous cache$^*$  & 22.95 & 23.59 \da{2.28} \\
  {\pfix}  $\ourlocal$ & 24.45 & 25.60 \dagray{0.27}  \\
  {\pfix}  $\ourlong$ & \tf{21.76} & \tf{22.66} {\dagray{3.21}}\\
    \bottomrule
    \end{tabular}
    }
    \vspace{-0.5em}
    \caption{Performance on the {\wiki} dataset (150M models, $L = 150$). $\ourlong$ uses a long-term memory with 15,000 tokens. 
    Transformer-XL: \cite{dai2019transformer}, Compressive Transformer: \cite{rae2019compressive}, $\infty$-former: \cite{martins2022infty}, LaMemo: \cite{ji2022lamemo}.
    $^*$: cache adapted to long-term memory.}
    \label{tab:wiki_150}
\end{table}

%% file: tables/enwik8.tex
\begin{table}[t]
    \centering
    \resizebox{0.98\columnwidth}{!}{
    \begin{tabular}{llll}
    \toprule
    \textbf{Model} & \textbf{\#Params} & \textbf{Dev} ($\downarrow$) & \textbf{Test}  ($\downarrow$) \\
    \midrule
    T12 & 44M & - & 1.11 \\
    Transformer-XL & 41M & - & 1.06 \\
    Adapt-Span & 39M & 1.04 & 1.02 \\
    Longformer & 41M & 1.02 & 1.00 \\
    Expire-Span & 38M & - & \tf{0.99} \\
    \midrule
    Transformer ($L =$ 512) & 38M & 1.18 & 1.16 \\
    \tableindent\tableindent + continuous cache$^*$  & 38M & 1.16 & 1.17 \ua{0.01}\\
   \tableindent\tableindent  {\pfix} $\ourlocal$ & 38M & 1.14 & 1.12 \dagray{0.04} \\
   \tableindent\tableindent  {\pfix} $\ourlong$ & 38M & \tf{1.08} & \tf{1.05} {\dagray{0.11}} \\
    \midrule
    SRU++ ($L$ = 512) & 42M & 1.05 & 1.03 \\
    \tableindent\tableindent {\pfix} $\ourlong$ & 42M & \tf{1.03} & \tf{1.01} {\dagray{0.02}} \\
    SRU++ ($L$ = 2,048) & 42M & 1.02 & 0.99 \\
    \tableindent\tableindent {\pfix} $\ourlong$ & 42M & \tf{1.00} & \tf{0.98} {\dagray{0.01}}\\
    \bottomrule
    \end{tabular}
    }
    \vspace{-0.5em}
    \caption{Performance on the {\enwik} dataset. $\ourlong$ achieves the best results by using a long-term memory of a size 24,576.
    T12: \cite{al2019character}, Transformer-XL: \cite{dai2019transformer}, Adapt-Span: \cite{sukhbaatar2019adaptive},  Longformer: \cite{beltagy2020longformer}, Expire-Span: \cite{sukhbaatar2021not}, SRU++: \cite{lei2021srupp}.
    $^*$: cache adapted to long-term memory.
    }
    \vspace{-0.5em}
    \label{tab:enwik8}
\end{table}

%% file: tables/bookcorpus.tex
\begin{table}[t]
    \centering
    \resizebox{1.0\columnwidth}{!}{
    \begin{tabular}{llll}
    \toprule
    \textbf{Model} & \textbf{$\externalm$} & \textbf{Dev} ($\downarrow$) & \textbf{Test}  ($\downarrow$) \\
    \midrule
    Transformer & - & 62.72 & 53.98 \\
    {\pfix} $\ourlocal$ & - & 59.39 & 49.25 \\
    {\pfix} $\ourlong$ & - & 49.21 & \tf{39.50} \\
    \midrule
    kNN-LM + cont. cache & \textsc{Wiki} & 53.27 & 43.24 \\
    {\pfix} $\ourext$ & \textsc{Wiki} & 47.00 & 37.70 \\
    kNN-LM + cont. cache & \textsc{Books} & 42.12 & 32.87 \\
    {\pfix} $\ourext$ & \textsc{Books} & 36.97 & \tf{27.84} \\
    \bottomrule
    \end{tabular}
    }
    \vspace{-0.5em}
    \caption{Domain-adaptation performance on the {\book} dataset. 
    All models are trained on {\wiki} and evaluated on {\book} without re-training or fine-tuning and we consider using {\wiki} and {\book} to build the external datastore respectively.  We use a long-term memory of a size 49,152 for $\ourlong$, $\ourext$, and continuous cache in this experiment.
    }
    \label{tab:bookcorpus}
\end{table}

%% file: tables/iwslt.tex
\begin{table}[t]
    \centering
    \resizebox{0.67\columnwidth}{!}{
    \begin{tabular}{ll}
    \toprule
    \textbf{Model} & \textbf{BLEU} ($\uparrow$) \\
    \midrule
    Transformer enc-dec & 32.58 \\
    kNN-MT & 33.15~{\uagray{0.57}} \\
    {\pfix} ${\ourmt}$ & \tf{33.73}~{\uagray{1.15}} \\
    \bottomrule
    \end{tabular}
    }
    \vspace{-0.5em}
    \caption{Results on the {\iwslt} De-En test set. We adapt {\ours} to the machine translation task. We use a beam size of 4 during evaluation.}
\vspace{-0.5em}
    \label{tab:iwslt}
\end{table}

%% file: sections/6-analysis.tex
\vspace{-0.5em}
\section{Analysis}
\label{sec:analysis}

\vspace{-0.3em}

We conduct ablation studies and analysis to further understand individual components of our approach.
Due to the limited computation budget, some experiments on {\wiki} are conducted with a small 7M Transformer model (8 layers, hidden dimension 128) in this section and the trends are generally similar for smaller models (see Appendix~\ref{sec:model_config} and Appendix~\ref{sec:wiki_small} for details).

\input{tables/batching}

\vspace{-0.5em}
\paragraph{Memory construction}
We first study how different data batching and memory construction strategies affect the performance when different testing memories are used. We compare our three models ({$\ourlocal$}, $\ourlong$, $\ourext$) in Table~\ref{tab:batching}.  This ablation study clearly shows that packing consecutive segments and segments with high BM25 scores in the same training batch and constructing memories properly can improve the performance when the long-range and external memories are used. 
This demonstrates the importance of closing the gap between training and inference.

\begin{figure}[t]
\centering
\includegraphics[width=0.95\linewidth]{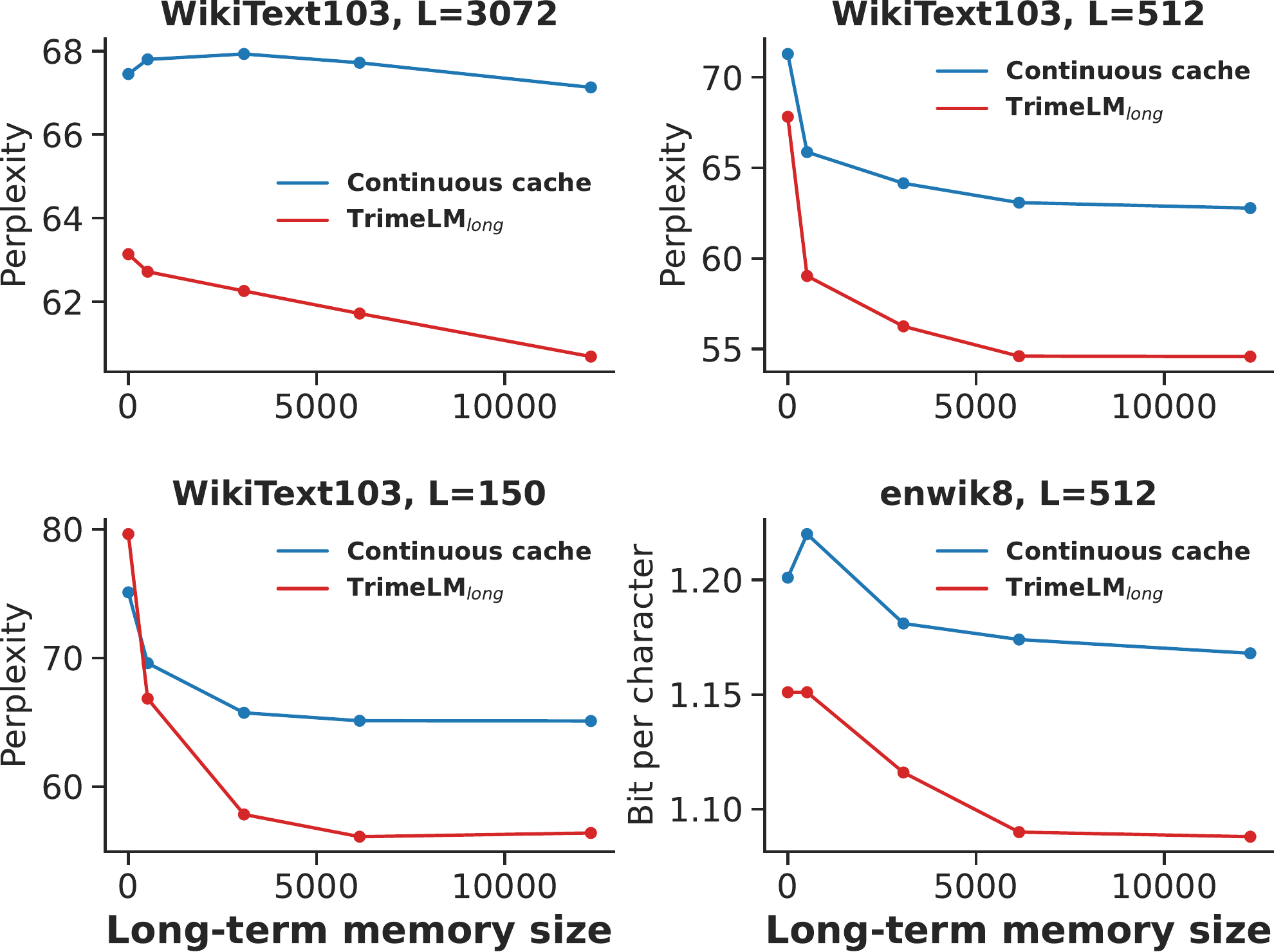}
\vspace{-0.5em}
\caption{
Performance of {$\ourlong$} (with different segment lengths $L$) and continuous cache (adapted to long-term memory) on the {\wiki} (7M models) and {\enwik} development sets. 
}
\label{fig:b_size}
\end{figure}

\vspace{-0.5em}
\paragraph{Leveraging long-range contexts}
We study if our model is able to handle large long-term memory.
As Figure~\ref{fig:b_size} shows, our model is able to effectively handle long-range context (more than 10k tokens), which goes beyond typical attention context.
Compared to continuous cache~\cite{grave2016improving,grave2017unbounded}, the improvement of our approach becomes larger when more long-term memory is incorporated. This suggests that our model is able to leverage long-range context much more effectively.

\vspace{-0.5em}
\paragraph{Additional analysis}
We conduct more ablation studies and analysis in Appendix~\ref{sec:add_analysis}. We summarize them as follows.
(1) Our ablation studies show using BM25 batching method and enabling back-propagation to update memory representations are important for our approach (Table~\ref{tab:ablation}).
(2) $\ourlocal$ is able to leverage local memory effectively to improve performance with different segment lengths $L$ (Table~\ref{tab:local_mem}).
(3) $\ourext$ outperforms kNN-LM in terms of top-K retrieval accuracy given the external memory set (Table~\ref{tab:retrieval}). 
(4) We study the perplexity of tokens in different frequency groups and find that $\ourlocal$ and $\ourlong$ achieve larger improvements on rare words while $\ourext$ improves results across the board (Table~\ref{tab:bucket_ppl}).

%% file: tables/batching.tex
\begin{table}[t]
    \centering
    \resizebox{0.95\columnwidth}{!}{

    \begin{tabular}{lccc}
    \toprule
    \tf{Method} & \multicolumn{3}{c}{\tf{Test memory}} \\
    \midrule
     &  &  {$\localm$}, & {$\localm$}, \\
     & {$\localm$} & {$\longm$} & {$\longm$}, {$\externalm$}\\
    \midrule
    $\ourlocal$ & \tf{17.10} & 17.17 & 17.40 \\
    $\ourlong$ & 17.12 & \tf{17.01} & 16.48 \\
    $\ourext$ & 17.99 & 17.80 & \tf{15.51} \\
    \bottomrule
    \end{tabular}
    }
    \vspace{-0.5em}
    \caption{Evaluating our three models (w/ different training methods) on different sets of testing memories. The results are based on the development set of {\wiki} (247M models, $L=$ 3,072).}
    \label{tab:batching}
\end{table}

%% file: sections/2-related.tex
\vspace{-0.2em}
\section{Related Work}
\label{sec:related}

\vspace{-0.5em}
\paragraph{Memory-augmented language models} We have discussed continuous cache, kNN-LM and models that leverage representations from long-range context in the previous sections.  \citet{yogatama2021adaptive} also aim to combine several types of memories by learning an adaptive gating function; however, their external memory uses a pre-trained vanilla language model. \citet{borgeaud2021improving} demonstrate a remarkable performance by augmenting LMs with an external datastore of trillion of tokens and their datastore is built based on chunks of text using off-the-shelf BERT embeddings~\cite{devlin2018bert}.
Our approach differs from prior works in the following aspects: (1) we update the memory representations through back-propagation from the end loss; (2) our model does not modify the base architecture; (3) we consider different types of memories in a unified framework.
GNN-LM~\cite{meng2022gnn} augments LMs with a graph neural network to aggregate information of retrieved items from external memory, which makes an orthogonal contribution to our paper.

\vspace{-0.5em}
\paragraph{Transformers for long inputs}
A large body of research has investigated how to scale self-attention mechanism to long contexts, either through sparse attention~\cite{liu2018generating,child2019generating,beltagy2020longformer, zaheer2020big} or sub-quadratic-time attention~\cite{wang2020linformer,choromanski2020rethinking,peng2021random,katharopoulos2020transformers}. See \citet{tay2020efficient} for a comprehensive survey of efficient Transformers.
Our approach is orthogonal, as we only change the training objective and data batching to enable models to use large contexts during inference. 

\vspace{-0.5em}
\paragraph{Memory-augmented models for downstream tasks}
Prior works have also improved models for downstream tasks with a retrieval component, such as question answering~\cite{kumar16ask,de2019episodic,karpukhin2020dense,guu2020realm,zemlyanskiy2021readtwice,de2022mention,chen2022augmenting,izacard2020leveraging,singh2021end}, dialogue~\cite{fan2021augmenting}, and other knowledge-intensive NLP tasks~\cite{lewis2020retrieval,petroni2021kilt}.
Notably, recent works~\cite{de2022mention,chen2022augmenting} explore a similar idea for question answering and leverage in-batch memories to train memory representations for entity mentions or QA pairs, which are further incorporated into Transformers at a second stage.

%% file: sections/7-conclusion.tex
\vspace{-0.3em}
\section{Conclusion}
\label{sec:conclusion}

\vspace{-0.3em}
In this work, we propose {\ours}, a training approach for language modeling. 
We present three model instantiations $\ourlocal$, $\ourlong$, $\ourext$: Through carefully-designed data batching and memory construction during training, we show that our models can leverage long-range contexts and external memory effectively at testing time. Our approach adds little computational overhead and does not modify model architectures, making it compatible with other neural models and techniques.  For future work, we are interested in training {\ours} with large language models and other text generation tasks.

%% file: appendix.tex
\appendix

\clearpage

\section{Inference Method}
\label{sec:inference}
\paragraph{Testing objective}
Formally speaking, our testing objective is basically the same as the training objective (Eq.~\ref{equ:loss}):
\begin{align}
\begin{split}
\label{equ:inf_loss}
    & P(w \mid c)  \propto \exp(E^{\top}_w f_{\theta}(c)) + \\
    & \sum_{(c_j, x_j) \in \testm: x_j = w} \exp(\frac{\simfunc(g_{\theta}(c), g_{\theta}(c_j)) }{\tau}),
\end{split}
\end{align}
except that we take $\testm$ as a combination of $\localm$, $\longm$ and $\externalm$. 
As $\externalm$ can be very large, we approximate it by retrieving the top-K closest terms to $g_\theta(c)$.
Formally, $\testm$ of three instantiations of $\ours$ is constructed as follows,
\begin{equation}
\resizebox{1.05\columnwidth}{!}{
$\testm = \begin{cases}
\localm\quad &{\scriptstyle(\ourlocal)} \\
\localm \cup \longm\quad &{\scriptstyle(\ourlong)} \\
\localm \cup \longm \cup \text{kNN}(\externalm, g_\theta(c))\quad &{\scriptstyle(\ourext)}
\end{cases} $
}
\end{equation}
where $\text{kNN}(\externalm, g_\theta(c))$ returns the top-K closest terms to $g_\theta(c)$ in the memory set $\externalm$.
Additionally, because $\testm$ may be different from the training memories, we tune a temperature term $\tau$ to adjust the weight of the memory component when calibrating the distribution, based on the development set.

\paragraph{Linear interpolation when using $\externalm$}
\label{sec:interpolation}
We find that when a large set of external memory $\externalm$ is considered during inference, the performance can be improved by calibrating a separated distribution over the memory and interpolating the output distribution and the memory distribution, similarly to kNN-LM~\cite{khandelwal2020generalization}.
We think this is because the distribution of the similarity values has been significantly shifted during inference, while the relative ranking preserves.
As a result, having values from two different distributions in one softmax normalization is sub-optimal compared to computing two separated probabilities and interpolating them. 

Thus, we apply an additional linear interpolation to our output probability distribution. Specifically, we first use Eq.~\ref{equ:inf_loss} to compute the distribution $P(w \mid c)$. 
Then, we compute a probability distribution over the tokens in memory $P'(w \mid c)$ as follow,
\begin{align}
\resizebox{1.05\columnwidth}{!}{$
    P'(w \mid c)  \propto \sum_{(c_j, x_j) \in \testm: x_j = w} \exp(\frac{\simfunc(g_{\theta}(c), g_{\theta}(c_j)) }{\tau'}).$
}
\end{align}
We linearly interpolate these two probability distributions with a coefficient $\lambda$ and get the final output $P_\text{final}(w \mid c)$:
\begin{align}
    P_\text{final}(w \mid c) = (1 - \lambda) P(w \mid c) + \lambda P'(w \mid c).
\end{align}
We tune the temperature terms and $\lambda$ on the development set.

\RestyleAlgo{ruled}
\SetKwComment{Comment}{/* }{ */}
\SetKw{Break}{break}

\begin{algorithm}[hbt!]
\caption{Packing segments using BM25 scores. $\texttt{SimSeg}(I, c, k)$ returns the top-$k$ most similar segments to $c$ in the BM25 indexer $I$. ($k=20$ when packing segments in our experiments.)}
\label{bm25_algo}
\KwData{training segments $S=\{s_1, \dots, s_{|S|}\}$}
\SetKwInput{KwData}{BM25 Indexer}
\KwData{$I$}
\SetKwInput{KwData}{Hyper-parameters}
\KwData{$k$, batch size $B$}
\SetKwInput{KwResult}{Output}
\KwResult{training batches $T$}
$l \gets $ list$()$\;
$c \gets $ \texttt{None}\;
\While{$|S| \neq 0$}{
    \If{$c$ is \texttt{None}}{
        $c \gets \texttt{random\_sample}(S)$\;
    }
    $l.\texttt{append}(c)$\;
    $S.\texttt{remove}(c)$\;
    
    $n \gets \texttt{None}$\;
    \For{$c'$ in $\texttt{SimSeg}(I, c, k)$}{
        \If{$c'$ in $S$}{$n \gets c'$\; \Break\;}
    }
    $c \gets n$\;
}
$T \gets \{[l_1, \dots, l_{B}], [l_{B+1}, \dots, l_{2B}], \dots \}$\;
\SetKw{Break}{return}
\Break $T$\;
\end{algorithm}

\input{tables/dataset}

\input{tables/model_config}

\input{tables/wiki_small}

\section{Packing Segments Using BM25 Scores}
\label{sec:bm25_batching}
In \S\ref{sec:external}, we construct training memories $\trainm$ by packing segments that have large lexical overlap into the same batch using BM25~\cite{robertson2009probabilistic}. 
Algorithm~\ref{bm25_algo} shows the process to pack segments into training batches. We start with a single segment and repeatedly add segments with highest BM25 scores into the same batch.

\input{tables/ablation_ext}

\input{tables/local_mem}

\input{tables/retrieval}

\input{tables/bucket_ppl}
\input{tables/disable}

\section{Dataset Statistics and Tasks}
\label{sec:dataset}
We evaluate our approach on three benchmarks: {\wiki}, {\enwik}, and {\iwslt}. We also evaluate our approach on {\book} for domain adaptation (Appendix~\ref{sec:domain}). Table~\ref{tab:datasets} shows the statistics.

\tf{{\wiki}}~\cite{merity2016pointer} is a word-level language modeling dataset consisting of 103M training tokens. 
Following standard practice, we use adaptive softmax and adaptive token embeddings~\cite{baevski2018adaptive} in our model and report perplexity.
In order to better compare with previous work, we evaluate on two model configurations---one uses a 247M Transformer model and a segment length $L=3,072$ following \citet{baevski2018adaptive,khandelwal2020generalization} and another one uses a 150M Transformer model with segment length $L=150$ following \citet{dai2019transformer}.
More details are provided in Appendix~\ref{sec:model_config}.

\tf{{\enwik}}~\cite{enwik8} is a character-level language modeling dataset that contains a total of 100M characters. Following previous work, we report bit-per-character (bpc) on this dataset.  
We use a 12-layer Transformer model with a hidden dimension $512$ and segment length $L = 512$.

We also evaluate the \tf{\iwslt} \textsc{De$\to$En} machine translation task, which consists of 170K translation pairs. 
Following \citet{khandelwal2021nearest}, we build an external memory by taking all the translation contexts and the corresponding target token $((x, y_{<t}), y_t)$ on the training set.
We use the output representation as $f((x, y_{<t}))$ and the input representation of last FFN layer as $g((x, y_{<t}))$ to compute the loss.
Similarly, we use BM25 to batch training data -- we encourage two target sentences with a high BM25 score to be in the same training batch (see Algorithm~\ref{bm25_algo}).
We use the default model configuration in the Fairseq library~\cite{ott2019fairseq}, and sacrebleu~\cite{post-2018-call} to compute BLEU scores~\cite{papineni2002bleu}.

We evaluate our approach for domain adaptation on the \tf{{\book}} dataset~\cite{zhu2015aligning}, which is a word-level language modeling dataset. The complete {\book} dataset consists of 0.7B tokens. We build our own train/dev/test splits which consist of 100M/250K/250K tokens respectively. The train set is only used to build external memory. On this dataset, we evaluate the models trained on {\wiki} to study how our approach can adapt to new domain without re-training or fine-tuning. The model we used on this dataset is the 247M Transformer model with a segment length $L=$ 3,072.

\section{Model Configurations and Hyperparameters}
\label{sec:model_config}

Table~\ref{tab:model_config} shows the model configurations and hyperparameters that we used in our experiments.
Following \citet{baevski2018adaptive}, during training, we train the model with fixed-length segments; during evaluation, we evaluate on the tokens at the end of the segment (i.e., an evaluation segment can overlap with others).

When evaluating with large external memory, we always retrieve top-$K$ ($K = $1,024) context-target pairs for language modeling. For machine translation, we tune $K = \{1, 2, 4, 8, 16, 32, 64\}$ following \citet{zheng2021adaptive}.

\section{Applying {$\ourlong$} to SRU++}
\label{sec:apply_to_sru}
We apply our approach to SRU++~\cite{lei2021srupp} and we believe our approach is also compatible with other architectures such as Transformer-XL~\cite{dai2019transformer}.
SRU++ is a language model which combines recurrent units and the attention mechanism.
SRU++ use hidden representations from the previous segment at attention layers to incorporate long-range contexts, similarly to \citet{dai2019transformer}.

To apply our approach to SRU++, we follow their data-batching method as it is required due to the recurrence of the model architecture. We construct the training memory using all the contexts in the current segment (i.e., local memory) and all contexts in the previous segment (i.e., long memory). 
Note that the memory representations from the previous segment will be stale, thus we do not back-propagate to that part. During training, we update the model with 400K steps and a batch size of $16$. For other hyper-parameters and the optimizer, we follow the default ones in their implementation.

During inference, we can use more contexts to construct memory. We train with different segment lengths, i.e., $L=512$ or $L=2048$. For the model trained with $L=512$, it can leverage a long-term memory of a size 6,144 during inference; for the model trained with $L=2048$, it can leverage a long-term memory of a size 12,228.

\section{Performance of the 7M model on {\wiki}}
\label{sec:wiki_small}

We conduct our ablation studies and analyses in \S\ref{sec:analysis} with an 8-layer Transformer model due to the limited computation budget.
The model consists of 7M parameters, 8 layers and 4 heads in each layer. The embedding dimension is 128 and the intermediate dimension of FFN is 512. The model takes a segment of 3072 tokens as input.
We compare our approach with baselines on this model architecture.  As shown in Table~\ref{tab:wiki_small}, our approach improves over the baselines by a large margin. This shows that modeling memory explicitly is essential when the model capacity is limited.

\section{Additional Analysis}
\label{sec:add_analysis}

\paragraph{Ablation study on {$\ourext$}}
We study the importance of packing segments with high BM25 scores in the same training batch, as well as the effectiveness of enabling back-propagation to memory representations during training. As shown in Table~\ref{tab:ablation}, when we random batch training segments (instead of using BM25 scores), the perplexity increases to 45.71 ($+4.21$). Also, enabling back-propagation to memory is crucial for our approach --- the performance is much worse if we disable it.

\paragraph{Effectiveness of using local memory}
\label{sec:eff_local}

We study the effectiveness of our model ${\ourlocal}$ that uses only local memory with different segment lengths $L$. As shown in Table~\ref{tab:local_mem}, our model significantly outperforms the baselines in all the settings. This suggests that our model can leverage local memory very effectively to improve performance.

\paragraph{Retrieval performance on external memory}
\label{sec:ret_perf}
When external memory is used in our experiments, we perform nearest-neighbor search over the entire memory set $\externalm$ to retrieve the top $K$ keys (we use $K=1024$).
Table~\ref{tab:retrieval} compares the retrieval accuracy of our approach and kNN-LM~\cite{khandelwal2020generalization} for different $K$.
Our approach outperforms kNN-LM in terms of retrieval results; this explains how our final perplexity surpasses kNN-LM when incorporating external memory.

\paragraph{Perplexity breakdown for different frequencies}
\label{sec:bucket_ppl}
We aim to understand which type of memories improves perplexity of tokens in different frequency groups.
We group tokens into 5 buckets according to their frequency on the development set.
Table~\ref{tab:bucket_ppl} shows the results for different models.
$\ourlocal$ and $\ourlong$ improve the perplexity of rare words (i.e., frequency $\leq$ 1k) while achieving similar or slightly worse results for frequent words compared to the Transformer baseline. $\ourext$ improves perplexity in all the buckets. 
Interestingly, kNN-LM with continuous cache does not perform significantly better compared to $\ourlocal$ and $\ourlong$ although these two models do not use external memory.
This suggests that jointly training memory representations and the language model particularly help improve the performance of rare words.

\section{Tuning $p$ for training with external memory}
\label{sec:tuning_p}
When training the model with local and external memory, to avoid the model to only relies on high-quality local memory, we disable the local memory with a probability of $p$. Here we study how $p$ will affect the final performance of our model. The results of using different $p$ are shown in Table~\ref{tab:disable}.
We find that when $p=0$, the model performs poorly with external memory as the model learns to only leverage local memory and ignores external memory during training. By increasing $p$, this issue is mitigated. We set $p=0.9$ in our main experiments.

%% file: tables/dataset.tex
\begin{table}[h]
    \centering
    \resizebox{1.0\columnwidth}{!}{
    \begin{tabular}{llllll}
    \toprule
    & \tf{Train} & \tf{Dev} & \tf{Test} & $|\mathcal{V}|$ & len \\
    \midrule
    {\wiki} & 110M & 0.2M & 0.3M & 270K & 3.6K\\
    {\enwik} & 94M & 5.2M & 5.2M & 256 & - \\
    {\book} & 100M & 250K & 250K & 270K & 90K \\
    \midrule
    {\iwslt} & 160K & 7K & 6K & 16K & 25 \\
    \bottomrule
    \end{tabular}
    }
    \caption{Statistics of the four datasets used in our paper. {\wiki} and {\book} are a word-level LM task and {\enwik} is a character-level language modeling task, and {\iwslt} is a German-English machine translation task. len:  denotes the average number of tokens over training examples in each dataset.
    When evaluating models on {\book} without re-training, we use the {\wiki}'s vocabulary.
    {\iwslt} is a sentence-level task, so incorporating long-range context will not help. }
    \label{tab:datasets}
\end{table}

%% file: tables/model_config.tex
\begin{table*}[h]
\centering
    \resizebox{1.6\columnwidth}{!}{
\begin{tabular}{l|ccc|c|c}
\toprule
\tf{Dataset}          & \multicolumn{3}{c|}{\wiki} & {\enwik} & {\iwslt} \\
\midrule
\tf{Model} & & & & & \\
~~~\#Params       & 247M      & 150M     & 7M       & 38M    & 39M      \\
~~~\#Layers         & 16        & 16       & 8        & 12     & 6+6      \\
~~~Hidden dimension    & 1024      & 410      & 128      & 512    & 512      \\
~~~FFN intermediate dimension    & 4096      & 2100     & 512      & 2048   & 1024     \\
~~~Adaptive softmax?        & yes       & yes      & yes      & no     & no       \\
\midrule
\tf{Training} & & & & &  \\
~~~Segment length      & 3072      & 150      & 3072     & 512    & -        \\
~~~\#Tokens per update & 73728 & 36000 & 24576 & 49152 & 16384 \\
~~~Gradient accumulation & 3 & 4 & 1 & 4 & 1 \\
~~~Batch size per update & 24 & 240 & 8 & 96 & - \\
~~~\#Consecutive segments & 4 & 60 & 8 & 24 & - \\
~~~\#In-batch memories & 24576 & 9000 & 24576 & 12288 & - \\
\midrule
\tf{Evaluation} & & & & &  \\
~~~Segment length & 512       & 64       & 512      & 80     & -       \\
~~~\#Optimal long-term memories & 12288 & 15000 & 12288 & 24576 & -\\
\midrule
\tf{Optimizer and scheduler} & & & & & \\
~~~Optimizer type & nag & adam & adam & adam & adam\\
~~~Learning rate & 1.0 & 5e-4 & 5e-4 & 2.5e-4 & 5e-4 \\
~~~Grad crop norm & 0.1 & 0.0 & 0.0 & 0.0 & 0.0\\
~~~Update steps & 286000 & 200000 & 200000 & 400000 & 170000\\
~~~Scheduler type & cosine & inverse\_sqrt & inverse\_sqrt & cosine & cosine\\
~~~Linear warmup steps & 16000 & 8000 & 8000 & 0 & 4000\\
\bottomrule
\end{tabular}
}
\caption{Model configurations and hyperparameters in our experiments.  
}
\label{tab:model_config}
\end{table*}

%% file: tables/wiki_small.tex
\begin{table}[t]
    \centering
    \resizebox{1.0\columnwidth}{!}{
    \begin{tabular}{lll}
    \toprule
    \textbf{Model} & \textbf{Dev} ($\downarrow$) & \textbf{Test} ($\downarrow$) \\
    \midrule
    Transformer & 82.68 & 83.66 \\
    ~~~~ + continuous cache & 67.45 & 68.08 \\
    kNN-LM & 51.88 & 52.24 \\
    ~~~~ + continuous cache & 45.15 & 45.82 \\
    {\pfix} $\ourlocal$ & 54.78  & 54.69 \dagray{28.97} \\
    {\pfix} $\ourlong$ & 60.71  & 60.10 \dagray{23.56} \\
    {\pfix} $\ourext$ & \tf{41.50} & \tf{42.36} \tf{\dagray{41.30}} \\
    \bottomrule
    \end{tabular}
    }
    \vspace{-0.5em}
    \caption{Performance of the 7M Transformer models on the {\wiki} dataset. 
    }
    \label{tab:wiki_small}
\end{table}

%% file: tables/ablation_ext.tex
\begin{table}[t]
    \centering
    \begin{tabular}{ll}
    \toprule
    \textbf{Model} & \textbf{Dev} ($\downarrow$) \\
    \midrule
    $\ourext$ & 41.50 \\
    ~~~~ w/o BM25 batching & 45.71 \\
    ~~~~ w/o back-prop to memory & 45.15 \\
    \bottomrule
    \end{tabular}
    \vspace{-0.5em}
    \caption{Ablation studies of using BM25 batching and enabling back-propagation to memory representations during training. The numbers are on the {\wiki} development set (7M models).
    }
    \label{tab:ablation}
\end{table}

%% file: tables/local_mem.tex
\begin{table}[h]
    \centering
    \resizebox{1.0\columnwidth}{!}{
    \begin{tabular}{lccc}
    \toprule
    \textbf{Model} & $L=3072$ & $L=512$ & $L=150$ \\
    \midrule
    Transformer & 82.70 & 81.15 & 81.40 \\
    ~~ + cont. cache & 67.45 & 71.29 & 75.10 \\
    {\pfix} $\ourlocal$ & 54.89 & 63.22 & 71.82 \\
    \bottomrule
    \end{tabular}
    }
    \vspace{-0.5em}
    \caption{Performance on the {\wiki} development set (7M models). We vary the segment $L$ here to study the effectiveness of using local memory.}
    \label{tab:local_mem}
\end{table}

%% file: tables/retrieval.tex
\begin{table}[!h]
    \centering
    \resizebox{1.0\columnwidth}{!}{
    \begin{tabular}{l|cccc}
    \toprule
    \tf{Model} & Top-1 & Top-8 & Top-64 & Top-1024\\
    \midrule
    kNN-LM & 25.82 & 50.03 & 69.85 & 86.97\\
    {\pfix} $\ourext$ & \tf{27.64} & \tf{51.16} & \tf{70.43} & \tf{87.18} \\
    \bottomrule
    \end{tabular}
    }
    \vspace{-0.5em}
    \caption{Retrieval performance on external memory of our model (7M) and kNN-LM~\cite{khandelwal2020generalization} on the {\wiki} development set. We report top-$K$ retrieval accuracy ($K=1, 8, 64, 1024$).
    }
    \label{tab:retrieval}
\end{table}

%% file: tables/bucket_ppl.tex
\begin{table*}[t]
    \centering

        \resizebox{1.5\columnwidth}{!}{
    \begin{tabular}{l|cccccc}
    \toprule
       Frequency  & $>$ 10k & 1k-10k & 100-1k & 10-100 & $\leq$ 10 & avg \\
    \midrule
       Transformer & 3.35 & 4.11 & 13.63 & 30.46 & 240.39 & 18.04 \\
       kNN-LM + cont. cache & \tf{3.14} & 3.85 & 12.92 & 26.90 & 196.03 & 16.23 \\
      \midrule
    {\pfix} $\ourlocal$ & 3.47 & 4.15 & 13.57 & 28.05 & 198.33 & 17.10 \\
    {\pfix} $\ourlong$ & 3.43 & 4.13 & 13.62 & 27.89 & 194.89 & 17.01 \\
    {\pfix} $\ourext$ & {3.15} & \tf{3.84} & \tf{12.50} & \tf{25.41} & \tf{171.61} & \tf{15.51} \\
    \bottomrule
    \end{tabular}
    }

    \caption{Averaged perplexity in each frequency bucket on the {\wiki} development set (247M models).
    }
    \label{tab:bucket_ppl}
\end{table*}

%% file: tables/disable.tex
\begin{table}[t]
    \centering
    \resizebox{1.0\columnwidth}{!}{
    \begin{tabular}{l|ccccc}
    \toprule
    $p$ & 0.0 & 0.1 & 0.5 & 0.9 & 1.0 \\
    \midrule
    Perplexity & 54.33 & 49.85 & 45.08 & 41.50 & 41.86\\
    \bottomrule
    \end{tabular}
    }
    \caption{The performance of our model $\ourext$ on the development set ({\wiki}, 7M models). We disable the local memory with a probability of $p$  during training.}
    \label{tab:disable}
\end{table}